\batchmode
\makeatletter
\def\input@path{{\string"E:/Trabajo Angel/Mis articulos/Finished/PL Gaussian proccess classification/Accepted/\string"}}
\makeatother
\documentclass[english]{IEEEtran}
\usepackage[T1]{fontenc}
\usepackage[latin9]{inputenc}
\usepackage{array}
\usepackage{booktabs}
\usepackage{multirow}
\usepackage{amsmath}
\usepackage{amssymb}
\usepackage{graphicx}

\makeatletter

\providecommand{\tabularnewline}{\\}

\usepackage{cite} 
\usepackage[margin=8pt,font=footnotesize]{caption}
\allowdisplaybreaks 

\makeatother

\usepackage{babel}
\begin{document}

\title{Gaussian process classification using posterior linearisation}

\author{Ángel F. García-Fernández, Filip Tronarp, Simo Särkkä\thanks{A. F. García-Fernández is with the Department of Electrical Engineering and Electronics, University of Liverpool, Liverpool L69 3GJ, United Kingdom. He is also with the ARIES Research Center, Universidad Antonio de Nebrija, Madrid, Spain (email: angel.garcia-fernandez@liverpool.ac.uk). 
F. Tronarp and S. Särkkä are with the  Department of Electrical Engineering and Automation, Aalto University, 02150 Espoo, Finland (emails: \{filip.tronarp, simo.sarkka\}@aalto.fi).}}
\maketitle
\begin{abstract}
This paper proposes a new algorithm for Gaussian process classification
based on posterior linearisation (PL). In PL, a Gaussian approximation
to the posterior density is obtained iteratively using the best possible
linearisation of the conditional mean of the labels and accounting
for the linearisation error. PL has some theoretical advantages over
expectation propagation (EP): all calculated covariance matrices are
positive definite and there is a local convergence theorem. In experimental
data, PL has better performance than EP with the noisy threshold likelihood
and the parallel implementation of the algorithms.
\end{abstract}


\section{Introduction}

Classification is an important problem with a large number of applications,
for example, in handwriting and speech recognition, and medical diagnosis
\cite{Murphy_book12}. In (supervised) classification, a set of training
data points with their corresponding classes are available to learn
the underlying structure of the problem. Based on this information,
the objective is to infer the classes of new data points. This classification
problem can be posed using Gaussian processes (GPs) \cite{Rasmussen_book06,Perez-Cruz13,Kim06,Markov13,Xiao15,Morales-Alvarez18,Olmos10}.

In binary GP classification, it is assumed that there is a latent
function, distributed as a GP \cite{Rasmussen_book06}, whose value
at a certain data point is related to the probability that this data
point belongs to one class. The GP prior has some hyperparameters
that can be marginalised out \cite{Vanhatalo10}, or estimated by
maximising the log marginal likelihood \cite{Rasmussen_book06}. Then,
for the estimated hyperparameters, we compute the posterior distribution
over the latent function evaluated at the training data points, which
in turn allows us to predict the classes for new data points. 

The main difficulties in GP classification are the approximations
of the posterior and the log marginal likelihood. Markov chain Monte
Carlo algorithms \cite{Neal98} can provide very accurate approximations,
but they usually have a high computational burden. This is the reason
why there is interest in using computationally efficient approximations.
We proceed to review two of such approximations in the literature,
though other approaches exists \cite{Gibbs00,Opper00,Hensman15}. 

One possibility is to use the Laplace approximation \cite{Rasmussen_book06}.
A drawback of the Laplace approximation is that it cannot handle likelihood
functions in which the gradient is zero almost everywhere, such as
the noisy threshold likelihood \cite{Kim06}. 

Expectation propagation (EP) \cite{Minka01} can be used with all
likelihood functions and often outperforms Laplace approximation in
GP classification \cite{Kuss05,Kuss06,Villacampa07}. EP is an iterative
algorithm in which, at each iteration, a Gaussian approximation to
the likelihood for one data point is reapproximated. In order to do
this, we first remove the considered approximated likelihood from
the posterior approximation, which results in the ``cavity'' distribution.
Then, we use the true likelihood and the cavity distribution to provide
a new Gaussian approximation to the likelihood by performing moment
matching. EP has two drawbacks that are relevant to this paper: 1)
the cavity distributions can have negative-definite covariance matrices
with possibly large negative eigenvalues that are not due to numerical
errors \cite{Murphy_book12,Vehtari_arxiv18}, and 2) there is no convergence
proof in the literature that indicates conditions of convergence \cite{Rasmussen_book06}.
In order to deal with 1), simple, ad-hoc solutions are sometimes used,
for example, processing the likelihoods in a different order to see
if this resolves the issue, or arbitrarily setting the negative-definite
covariance to a predefined positive-definite matrix \cite{Vehtari_arxiv18}.
More robust EP algorithms, such as damped EP or double loop algorithms,
also require pragmatic solutions to avoid negative-definite covariance
matrices \cite{Jylanki11}. 

This paper proposes an algorithm for GP classification based on posterior
linearisation (PL) \cite{Angel15_c,Tronarp18}. In PL, we compute
a Gaussian approximation to the posterior distribution by linearising
the conditional mean of each label, in the region where the posterior
lies, and by setting the conditional covariance to a value that accounts
for the linearisation error. Importantly, the selection of the linearisation
parameters is done in an optimal way by minimising the mean square
error of the linearisation. This optimal linearisation is given by
statistical linear regression (SLR) \cite{Arasaratnam07} of the conditional
mean with respect to the posterior. 

In practice, PL is implemented by an iterative procedure, in which
we can process the likelihoods sequentially or in parallel. PL has
some advantages compared to EP: 1) PL always provides positive-definite
covariance matrices, so ad-hoc fixes are not necessary, and 2) there
is a local convergence proof \cite{Tronarp18} that indicates sufficient
local conditions of convergence. The proposed algorithm also includes
a method to approximate the marginal likelihood for estimating the
hyperparameters. In the analysed experimental data, EP and PL have
comparable performance in terms of classification errors except in
the parallel implementations for the noisy threshold likelihood, where
PL provides lower errors. 

\section{Problem formulation }

\label{sec:Problem-formulation}

We consider a set $\mathcal{D}=\left\{ \left(\mathbf{x}_{1},y_{1}\right),...,\left(\mathbf{x}_{n},y_{n}\right)\right\} $,
which contains $n$ data points $\mathbf{x}=\left(\mathbf{x}_{1},...,\mathbf{x}_{n}\right)$,
with $\mathbf{x}_{i}\in\mathbb{R}^{n_{x}}$, where $n_{x}$ is the
dimension of each data point, and their binary labels $\mathbf{y}=\left(y_{1},...,y_{n}\right)$,
with $y_{i}\in\left\{ -1,1\right\} $. In binary classification, given
this set, we are interested in predicting the binary labels $\mathbf{y}^{\star}=\left(y_{1}^{\star},...,y_{m}^{\star}\right)$
for $m$ new data points $\mathbf{x}^{\star}=\left(\mathbf{x}_{1}^{\star},...,\mathbf{x}_{m}^{\star}\right)$
\cite{Rasmussen_book06}. We proceed to explain how this problem is
formulated using GPs. 

It is assumed that the label $y_{i}$ of a data point $\mathbf{x}_{i}$
only depends on the value of a latent real-valued function evaluated
at $\mathbf{x}_{i}$, $f_{i}=f\left(\mathbf{x}_{i}\right)$. Then,
there are several widely-used models for the probability mass function
$p\left(y_{i}|f_{i}\right)$, for example, the probit, logit, and
noisy threshold, whose respective $p\left(y_{i}|f_{i}\right)$ are
 \cite{Kim06} 
\begin{align}
p\left(y_{i}|f_{i}\right) & =\Phi\left(y_{i}f_{i}\right),\label{eq:probit_likelihood}\\
p\left(y_{i}=1|f_{i}\right) & =1/\left(1+\exp\left(-f_{i}\right)\right),\label{eq:logit_likelihood}\\
p\left(y_{i}|f_{i}\right) & =\epsilon+\left(1-2\epsilon\right)H\left(y_{i}f_{i}\right),\label{eq:noisy_threshold_likelihood}
\end{align}
where $\Phi\left(z\right)=\int_{-\infty}^{z}\mathcal{N}\left(x;0,1\right)dx$,
$\mathcal{N}\left(\cdot;\overline{x},P\right)$ is the Gaussian density
with mean $\overline{x}$ and covariance $P$, $\epsilon\in\left(0,1\right)$,
and $H\left(z\right)=1$ if $z>0$ , and $H\left(z\right)=0$ otherwise. 

It is further assumed that function $f\left(\cdot\right)$ is distributed
according to a zero-mean Gaussian process with a given covariance
function $k_{\theta}\left(\cdot,\cdot\right)$, where $\theta\in\mathbb{R}^{n_{\theta}}$
is a vector that contains all the hyperparameters. As a result, the
prior density of $\mathbf{f}=\left[f_{1},...,f_{n}\right]^{\top}$
becomes
\begin{align}
p\left(\mathbf{f}|\mathbf{x},\theta\right) & =\mathcal{N}\left(\mathbf{f};\mathbf{0},\mathbf{K}\right),\label{eq:prior_f_x}
\end{align}
where $\mathbf{K}$ is an $n\times n$ covariance matrix such that
$\mathbf{K}{}_{ij}=k_{\theta}\left(\mathbf{x}_{i},\mathbf{x}_{j}\right)$.
The posterior of $\mathbf{f}$ is 
\begin{align}
p\left(\mathbf{f}|\mathcal{D};\theta\right) & =\left[\prod_{i=1}^{n}p\left(y_{i}|f_{i}\right)\right]p\left(\mathbf{f}|\mathbf{x};\theta\right)/p\left(\mathcal{D};\theta\right),\label{eq:posterior_f_x}
\end{align}
where the marginal likelihood is 
\begin{align}
p\left(\mathcal{D};\theta\right) & =\int\left[\prod_{i=1}^{n}p\left(y_{i}|f_{i}\right)\right]p\left(\mathbf{f}|\mathbf{x};\theta\right)\mathrm{d}\mathbf{f}.\label{eq:marginal_likelihood}
\end{align}

The hyperparameters $\theta$ are usually not known and are often
estimated by maximising (\ref{eq:marginal_likelihood}) \cite{Rasmussen_book06}.
Once we have estimated $\theta$, the posterior over $\mathbf{f}^{\star}=\mathbf{f}\left(\mathbf{x}^{\star}\right)=\left[f_{1}^{\star},...,f_{m}^{\star}\right]^{\top}$,
which considers new data points, is computed as 
\begin{align}
p\left(\mathbf{f}^{\star}|\mathcal{D},\mathbf{x}^{\star};\theta\right) & =\int p\left(\mathbf{f}^{\star}|\mathcal{D},\theta,\mathbf{x}^{\star},\mathbf{f}\right)p\left(\mathbf{f}|\mathcal{D};\theta\right)\mathrm{d}\mathbf{f}.\label{eq:posterior_f_x_ast}
\end{align}

Finally, all information about the class labels for the new data points
is given by the distribution of $\mathbf{y}^{\star}$ given $\mathcal{D}$,
$\mathbf{x}^{\star}$ and $\theta$, which can be written as 
\begin{align}
p\left(\mathbf{y}^{\star}|\mathcal{D},\mathbf{x}^{\star};\theta\right) & =\int\left[\prod_{i=1}^{m}p\left(y_{i}^{\star}|f_{i}^{\star}\right)\right]p\left(\mathbf{f}^{\star}|\mathcal{D},\mathbf{x}^{\star};\theta\right)\mathrm{d}\mathbf{f}^{\star}.\label{eq:predicted_measurement_ast}
\end{align}
Based on this distribution, we can predict the class labels, which
is our main objective, for example, by computing their expected value
\cite{Rasmussen_book06}. Unfortunately, none of the densities of
interest, (\ref{eq:posterior_f_x})-(\ref{eq:predicted_measurement_ast}),
has a closed-form expression, so approximations are necessary. As
in the Laplace and EP methods, in this paper, we consider Gaussian
approximations of (\ref{eq:posterior_f_x}) and (\ref{eq:posterior_f_x_ast}).

\subsection{Enabling approximation}

We obtain a Gaussian approximation to the posterior by approximating
the conditional mean $\mathrm{E}\left[y_{i}|f_{i}\right]$ as an affine
function and the conditional variance $\mathrm{C}\left[y_{i}|f_{i}\right]$
as a constant
\begin{align}
\mathrm{E}\left[y_{i}|f_{i}\right] & \approx A_{i}f_{i}+b_{i},\quad\mathrm{C}\left[y_{i}|f_{i}\right]\approx\Omega_{i},\label{eq:enabling_approximation}
\end{align}
where $A_{i}\in\mathbb{R}$, $b_{i}\in\mathbb{R}$, and $\Omega_{i}>0$,
and the conditional moments $\mathrm{E}\left[y_{i}|f_{i}\right]$
and $\mathrm{C}\left[y_{i}|f_{i}\right]$ are taken with respect to
$p\left(y_{i}|f_{i}\right)$, which is a discrete distribution. 

Under approximation (\ref{eq:enabling_approximation}), the linear
mean square error (LMMSE) estimate $\overline{\mathbf{u}}$ of $\mathbf{f}$
and its mean square error matrix $\mathbf{W}$ are available in closed-form
\cite{Anderson_book79}
\begin{align}
\overline{\mathbf{u}} & =\mathbf{K}\mathbf{A}^{\top}\left(\mathbf{A}\mathbf{K}\mathbf{A}^{\top}+\boldsymbol{\Omega}\right)^{-1}\left(\mathbf{y}-\mathbf{b}\right),\label{eq:posterior_mean}\\
\mathbf{W} & =\mathbf{K}-\mathbf{K}\mathbf{A}^{\top}\left(\mathbf{A}\mathbf{K}\mathbf{A}^{\top}+\boldsymbol{\Omega}\right)^{-1}\mathbf{A}\mathbf{K},\label{eq:posterior_covariance}
\end{align}
where $\mathbf{A}=\mathrm{diag}\left(\left[A_{1},...,A_{n}\right]\right)$,
$\mathbf{b}=\left[b_{1},...,b_{n}\right]^{\top}$ and $\boldsymbol{\Omega}=\mathrm{diag}\left(\left[\Omega_{1},...,\Omega_{n}\right]\right)$.
Then, the posterior (\ref{eq:posterior_f_x}) is approximated as Gaussian
with mean $\overline{\mathbf{u}}$ and covariance matrix $\mathbf{W}$,
which implies that the posterior of $\mathbf{f}^{\star}$, see (\ref{eq:posterior_f_x_ast}),
is Gaussian with mean $\overline{\mathbf{u}}^{\star}$ and covariance
matrix $\mathbf{W}^{\star}$ 
\begin{align}
\overline{\mathbf{u}}^{\star} & =\left(\mathbf{K}^{\star}\right)^{\top}\mathbf{A}^{\top}\left(\mathbf{A}\mathbf{K}\mathbf{A}^{\top}+\boldsymbol{\Omega}\right)^{-1}\left(\mathbf{y}-\mathbf{b}\right),\label{eq:posterior_mean-prediction}\\
\mathbf{W}^{\star} & =\mathbf{K}^{\star\star}-\left(\mathbf{K}^{\star}\right)^{\top}\mathbf{A}^{\top}\left(\mathbf{A}\mathbf{K}\mathbf{A}^{\top}+\boldsymbol{\Omega}\right)^{-1}\mathbf{A}\mathbf{K}^{\ast},\label{eq:posterior_covariance_prediction}
\end{align}
where $\mathbf{K}^{\star}$ and $\mathbf{K}^{\star\star}$ are $n\times m$
and $m\times m$ matrices such that $\mathbf{K}_{ij}^{\star}=k_{\theta}\left(\mathbf{x}_{i},\mathbf{x}_{j}^{\star}\right)$
and $\mathbf{K}_{ij}^{\star\ast}=k_{\theta}\left(\mathbf{x}_{i}^{\star},\mathbf{x}_{j}^{\star}\right)$. 

The quality of the approximations of the posterior moments (\ref{eq:posterior_mean})-(\ref{eq:posterior_covariance_prediction})
only depends on the choice of $A_{i},b_{i},\Omega_{i}$ for $i=1,...,n$,
so it is of utmost importance to select them properly. 

\section{Posterior linearisation of GPs}

\label{sec:Posterior-linearisation}

In this section, we first explain SLR using conditional moments in
Section \ref{subsec:Statistical-linear-regression}. Then, we explain
iterated posterior linearisation in Section \ref{subsec:Iterated-posterior-linearisation}.
We propose a method for approximating the marginal likelihood in Section
\ref{subsec:Marginal-likelihood-approximatio}. Finally, a discussion
of the algorithm is provided in Section \ref{subsec:Discussion}. 

\subsection{Statistical linear regression of conditional moments}

\label{subsec:Statistical-linear-regression}

With SLR, we can optimally linearise $\mathrm{E}\left[y_{i}|f_{i}\right]$
in a mean square error sense to make approximation (\ref{eq:enabling_approximation}).
In addition, SLR provides us with the linearisation error, which is
used to approximate $\mathrm{C}\left[y_{i}|f_{i}\right]$ as a constant,
as required in (\ref{eq:enabling_approximation}). The SLR linearisation
parameters are denoted as $\left(A_{i}^{+},b_{i}^{+},\Omega_{i}^{+}\right)$
and we proceed to explain how to obtain them. 

In SLR of random variables \cite{Tronarp18}, we are given a density
$p\left(\cdot\right)$ on variable $f_{i}$, whose first two moments
are $\overline{f_{i}}$ and $P_{i}$, and the conditional moments
$\mathrm{E}\left[y_{i}|f_{i}\right]$ and $\mathrm{C}\left[y_{i}|f_{i}\right]$,
which describe the relation between $y_{i}$ and $f_{i}$. Parameters
$A_{i}$ and $b_{i}$ are then selected by minimising the mean square
error over random variables $y_{i}$ and $f_{i}$
\begin{align}
\mathrm{E}_{f_{i},y_{i}}\left[\left(y_{i}-A_{i}f_{i}-b_{i}\right)^{2}\right] & =\mathrm{E}_{f_{i}}\left[\left(\mathrm{E}\left[y_{i}|f_{i}\right]-A_{i}f_{i}-b_{i}\right)^{2}\right]\nonumber \\
 & \quad+\mathrm{E}_{f_{i}}\left[\mathrm{C}\left[y_{i}|f_{i}\right]\right],\label{eq:mean_square_error_prev}
\end{align}
where we have highlighted the expectations that are taken with respect
to variables $f_{i}$ and $y_{i}$. Therefore, 
\begin{align*}
\left(A_{i}^{+},b_{i}^{+}\right) & =\underset{A_{i},b_{i}}{\arg\min}\,\mathrm{E}_{f_{i}}\left[\left(\mathrm{E}\left[y_{i}|f_{i}\right]-A_{i}f_{i}-b_{i}\right)^{2}\right].
\end{align*}

In SLR, the parameter $\Omega_{i}^{+}$ is the resulting mean square
error in (\ref{eq:mean_square_error_prev}) for the optimal values
$A_{i}^{+}$ and $b_{i}^{+}$ so
\begin{align*}
\Omega_{i}^{+} & =\mathrm{E}_{f_{i}}\left[\left(\mathrm{E}\left[y_{i}|f_{i}\right]-A_{i}^{+}f_{i}-b_{i}^{+}\right)^{2}\right]+\mathrm{E}_{f_{i}}\left[\mathrm{C}\left[y_{i}|f_{i}\right]\right].
\end{align*}
In other words, SLR makes the best affine fit of the conditional mean
$\mathrm{E}\left[y_{i}|f_{i}\right]$ in the region indicated by $p\left(\cdot\right)$,
the density of $f_{i}$, and sets $\mathrm{C}\left[y_{i}|f_{i}\right]$
as the corresponding mean square error. The resulting $\left(A_{i}^{+},b_{i}^{+},\Omega_{i}^{+}\right)$
is given by \cite{Tronarp18}
\begin{alignat}{1}
A_{i}^{+} & =\mathrm{C}_{f_{i}}\left[f_{i},\mathrm{E}\left[y_{i}|f_{i}\right]\right]/P_{i},\quad b_{i}^{+}=\mathrm{E}_{f_{i}}\left[\mathrm{E}\left[y_{i}|f_{i}\right]\right]-A_{i}^{+}\overline{f_{i}},\label{eq:A_plus}\\
\Omega_{i}^{+} & =\mathrm{C}_{f_{i}}\left[\mathrm{E}\left[y_{i}|f_{i}\right]\right]+\mathrm{E}_{f_{i}}\left[\mathrm{C}\left[y_{i}|f_{i}\right]\right]-\left(A_{i}^{+}\right)^{2}P_{i},\label{eq:Omega_plus}
\end{alignat}
where $\mathrm{C}_{f_{i}}\left[\cdot\right]$ denotes both the variance
and covariance with respect to $f_{i}$. We can then obtain $\left(A_{i}^{+},b_{i}^{+},\Omega_{i}^{+}\right)$
in terms of moments of $\mathrm{E}\left[y_{i}|f_{i}\right]$ and $\mathrm{C}\left[y_{i}|f_{i}\right]$
with respect to density $p\left(\cdot\right)$. Expressions of these
moments for the likelihoods (\ref{eq:probit_likelihood})-(\ref{eq:noisy_threshold_likelihood})
are given in the supplementary material.

\subsection{Iterated posterior linearisation}

\label{subsec:Iterated-posterior-linearisation}

This section explains how to use SLR to make approximations (\ref{eq:enabling_approximation})
in an optimal fashion. If we did not know the labels $y_{1},...,y_{n}$,
the best approximation of the conditional moments would be given by
SLR with respect to the prior, which is given by (\ref{eq:prior_f_x}),
as this density indicates the region where $f_{i}$ lies. However,
we know these labels, and the insight of posterior linearisation \cite{Angel15_c}
is that, given these labels, the best linearisation $\left(A_{i},b_{i}\right)$,
in a mean square error sense, and the resulting mean square error
$\Omega_{i}$ are given by SLR with respect to the posterior. 

Direct application of posterior linearisation is not useful as we
need to know the posterior to select $\left(\mathbf{A},\mathbf{b},\boldsymbol{\Omega}\right)$,
which is used to calculate the posterior. Nevertheless, posterior
linearisation can be approximated by using iterated SLRs, giving rise
to iterated posterior linearisation. That is, as we do not know the
posterior, we perform SLR with respect to the best available approximation
of the posterior. At the end of each iteration, we expect to obtain
an improved approximation of the posterior, which is used to compute
an improved SLR at the next iteration. The steps of the parallel version
of the algorithm are:
\begin{enumerate}
\item Set $j=1$ and $\overline{\mathbf{u}}^{j}=\mathbf{0},\,\mathbf{W}^{j}=\mathbf{K}$. 
\item For $i=1,...,n$, compute $\left(A_{i}^{j},b_{i}^{j},\Omega_{i}^{j}\right)$
using SLR with respect to the $i$th marginal of a Gaussian with moments
$\left(\overline{\mathbf{u}}^{j},\mathbf{W}^{j}\right)$, see (\ref{eq:A_plus})-(\ref{eq:Omega_plus}). 
\item With the current linearisation, $\left(A_{i}^{j},b_{i}^{j},\Omega_{i}^{j}\right)$
for $i=1,...,n$, compute the new posterior moments $\left(\overline{\mathbf{u}}^{j+1},\mathbf{W}^{j+1}\right)$
using (\ref{eq:posterior_mean}) and (\ref{eq:posterior_covariance}),
where only the values of $\left(\mathbf{A},\mathbf{b},\boldsymbol{\Omega}\right)$
change. 
\item Set $j=j+1$ and repeat from step 2 for a fixed number of iterations
or until some convergence criterion is met. 
\end{enumerate}
It is important to notice that in the aforementioned algorithm, we
first relinearise all likelihoods and then compute the posterior with
the current linearisation. This is beneficial from a computational
point of view because the linearisations can be done in parallel and
we only perform one update per iteration. Nevertheless, it is also
possible to recompute the posterior after relinearising a single likelihood.
In GP classification, $\mathbf{K}$ can be close to singular so we
have adapted the numerically stable implementations of Laplace and
EP algorithms in \cite{Rasmussen_book06} to our PL implementations.

\subsection{Marginal likelihood approximation}

\label{subsec:Marginal-likelihood-approximatio}

In this section, we propose the use of sigma-point/quadrature rules
to approximate the marginal likelihood, which is given by (\ref{eq:marginal_likelihood}).
Importantly, we select the quadrature points with respect to the posterior
approximation, as this density has its mass in the region where the
integrand is high. We therefore write (\ref{eq:marginal_likelihood})
as
\begin{align}
p\left(\mathcal{D};\theta\right) & =\hat{p}\left(\mathcal{D};\theta\right)\int\left[\prod_{i=1}^{n}\frac{p\left(y_{i}|f_{i}\right)}{\hat{p}\left(y_{i}|f_{i}\right)}\right]\nonumber \\
 & \quad\times\frac{\left[\prod_{i=1}^{n}\hat{p}\left(y_{i}|f_{i}\right)\right]\mathcal{N}\left(\mathbf{f};\mathbf{0},\mathbf{K}\right)}{\hat{p}\left(\mathcal{D};\theta\right)}d\mathbf{f},\label{eq:marginal_likelihood_manipulated}
\end{align}
where $\hat{p}\left(\mathcal{D};\theta\right)=\mathcal{N}\left(\mathbf{y};\mathbf{b},\mathbf{A}\mathbf{K}\mathbf{A}^{\top}+\boldsymbol{\Omega}\right)$
and $\hat{p}\left(y_{i}|f_{i}\right)=\mathcal{N}\left(y_{i};A_{i}f_{i}+b_{i},\Omega_{i}\right)$
$i=1,...,n$. Consequently, the marginal likelihood can be seen as
$\hat{p}\left(\mathcal{D};\theta\right)$ times a correction factor
that depends on the similarity between $p\left(y_{i}|f_{i}\right)$
and $\hat{p}\left(y_{i}|f_{i}\right)$ in the region indicated by
the posterior density. 

There are some drawbacks when integrating with respect to the joint
density of $\mathbf{f}$ using sigma-points/quadrature rules. First,
accurate and computationally efficient integration in high-dimensional
spaces is more difficult than in low-dimensional spaces. Second, sigma-point/quadrature
rules require the Cholesky decomposition of the covariance matrix
and this covariance can be ill-conditioned in GP classification \cite{Rasmussen_book06},
so it is not always possible to compute it. We therefore discard correlations
in the posterior for approximating the correction factor in (\ref{eq:marginal_likelihood_manipulated})
such that 
\begin{align}
p\left(\mathcal{D};\theta\right) & \approx\hat{p}\left(\mathcal{D};\theta\right)\prod_{i=1}^{n}\int\left[\frac{p\left(y_{i}|f_{i}\right)}{\hat{p}\left(y_{i}|f_{i}\right)}\right]\mathcal{N}\left(f_{i};\overline{u}_{i},W_{i}\right)\mathrm{d}f_{i},\label{eq:marginal_likelihood_approximated}
\end{align}
where $\overline{u}_{i}$ and $W_{i}$ represent the posterior mean
and variance of $f_{i}$, which are obtained from (\ref{eq:posterior_mean})
and (\ref{eq:posterior_covariance}). 

In short, in order to compute (\ref{eq:marginal_likelihood_approximated}),
we compute the posterior moments, $\overline{\mathbf{u}}$ and $\mathbf{W}$,
and the resulting SLR parameters $\left(\mathbf{A},\mathbf{b},\boldsymbol{\Omega}\right)$,
which are required in $\hat{p}\left(\mathcal{D};\theta\right)$ and
$\hat{p}\left(y_{i}|f_{i}\right)$. Accurate approximation of (\ref{eq:marginal_likelihood_manipulated})
is quite important as it is used to estimate the hyperparameters $\theta$.
Without an accurate estimation of $\theta$, the results of a GP classifier
are poor, as the GP does not model the training data properly. The
results in Section \ref{sec:Experimental-results} indicate that approximation
(\ref{eq:marginal_likelihood_approximated}) is accurate for classification
purposes. 

\subsection{Discussion}

\label{subsec:Discussion}

The iterated SLRs of PL can be done in parallel for each likelihood
and sequentially. We can also combine both types of linearisation
approaches, for example, by performing several updates sequentially
and then in parallel. Importantly, the main benefit of PL compared
to EP is that all involved densities in PL have positive-definite
covariance matrices. This is ensured by the fact that $\boldsymbol{\Omega}$
is positive definite by definition. Numerical inaccuracies could render
a negative-definite $\boldsymbol{\Omega}$ but this would be easy
to address, as its  eigenvalues would be close to zero. Another option
is to use square root solutions \cite{Arasaratnam08}. 

As EP, iterated PL is not ensured to converge in general. Nevertheless,
there is a local convergence proof, which is given in \cite[Thm. 2]{Tronarp18},
that states sufficient conditions for convergence.   

\section{Experimental results}

\label{sec:Experimental-results}

This section assesses Laplace, EP, and PL, in their parallel and sequential
forms, in six real-world data sets from \cite{Lichman:2013_uci_rep}.
One additional synthetic example that analyses a case where EP fails
is provided in the supplementary material. In particular, we consider
the data sets: breast cancer (9, 699), crab gender (6, 200), glass
chemical (9, 214), ionosphere (33, 351), thyroid (5,215) and housing
(13,506), where the number of attributes (dimension of data points)
and data points in each data set are given in parentheses. For the
last two data sets, the groups for binary classification are formed
as in \cite{Kim06}. Data points have been normalised to have zero
mean and an identity covariance matrix \cite{Kuss05}. We use ten-fold
cross-validation \cite{Rasmussen_book06} to compute the average classification
error. 

We use the covariance function \cite{Neal98} 
\begin{align*}
k_{\theta}\left(x_{i},x_{j}\right) & =\sigma_{1}^{2}\exp\left(-\frac{1}{2\ell^{2}}\left\Vert x_{i}-x_{j}\right\Vert ^{2}\right)+\sigma_{2}^{2}\delta\left[i-j\right],
\end{align*}
where $\delta\left[\cdot\right]$ denotes the Kronecker delta, $\sigma_{2}^{2}$
is set to 0.1, the hyperparameters $\theta=\left(\sigma_{1}^{2},\ell\right)$,
and $x_{i}\neq x_{j}$ for $i\neq j$. EP and Laplace have been implemented
as indicated in \cite{Rasmussen_book06}. 

We report results for all the likelihoods in (\ref{eq:probit_likelihood})-(\ref{eq:noisy_threshold_likelihood}).
The integrations that do not admit closed-form expressions are approximated
using a Gauss-Hermite quadrature of order 10. We first estimate the
hyperparameters $\theta$ by maximising (\ref{eq:marginal_likelihood})
using the BFGS Quasi-Newton algorithm implemented in Matlab function
\textit{fminunc}. The optimisation method is run on variable $\mathrm{ln}\left(\theta\right)$
with initial point $\left(\mathrm{ln}\,\sigma_{1}^{2},\mathrm{ln}\,\ell\right)=\left(\mathrm{ln}\left(10\right),\mathrm{ln}\left(1\right)\right)$.
Then, we approximate the posterior (\ref{eq:posterior_f_x}) as Gaussian
and, finally, we compute the expected value of (\ref{eq:predicted_measurement_ast})
to predict the label for each test data point. We consider 10 iterations
for EP and PL. If the variance of a likelihood approximation is negative
for EP, it is set to a small positive number to avoid negative-definite
covariance matrices, as suggested in \cite{Vehtari_arxiv18}.

The resulting average classification errors for each data set and
averaged over all data sets (Ave.) are shown in Table \ref{tab:Average-classification-errors},
where Prob., Log., and NT stand for probit, logit, and noisy threshold.
PEP and SEP refer to parallel and sequential EP, and PPL and SPL to
parallel and sequential PL. The cases where the classification error
is considerably high are bolded. In general, sequential algorithms
work better than parallel algorithms, though, as we will analyse,
with an increase in the computational burden. In fact, SEP and SPL
show comparable results. The highest differences among the methods
appear with the noisy threshold likelihood and the parallel implementations.
In this case, PPL works well but PEP provides high errors in four
out of six experiments. The reason is that EP returns negative-definite
covariance matrices for this likelihood and the fixes do not make
PEP attain a low error. Laplace does not work with this likelihood,
as the gradient of the likelihood is zero. 

For the logit model, PEP provides a high error in the cancer data
set. This is a numerical error that can be solved by using a Gauss-Hermite
quadrature rule of order 32, which increases the computational load.
In contrast, PPL is able to work well in this data set with the Gauss-Hermite
quadrature rule of order 10. The sequential processing of the data
for EP and PL work well in this data set. In the rest of the experiments,
EP, PL and Laplace provide comparable results, though there are slight
differences in the errors.  Importantly, parallel PL is more robust
than parallel EP, as errors are not markedly high in any of the data
sets. We think this is a relevant, practical advantage of PL over
EP.

Finally, we provide the average computational times of our Matlab
implementations (probit model and crab data set), as a indication
of their computational complexities, though running times depend on
the data set. With an Intel Core Xenon processor, we have: 0.4 s (Laplace),
0.7 s (PEP), 6.1 s (SEP), 0.8 s (PPL), and 5.5 s (SPL). Laplace is
the fastest algorithm. PEP is slightly faster than PPL, as its optimisation
requires a slightly fewer number of iterations to converge, although
PEP has robustness problems for the noisy threshold likelihood. Sequential
algorithms are slower, but they usually perform better than parallel
algorithms. In this case, SPL is marginally faster than SEP and both
methods have comparable performance. 

\begin{table}
\caption{\label{tab:Average-classification-errors}Average classification errors
for real data sets.}
\centering{}%
\begin{tabular}{>{\raggedright}p{0.28cm}>{\raggedright}p{0.28cm}lllllll}
Like. &
Alg. &
\multicolumn{1}{l}{Can.} &
\multicolumn{1}{l}{Crab} &
\multicolumn{1}{l}{Glass} &
\multicolumn{1}{l}{Ionos.} &
\multicolumn{1}{l}{Thy.} &
\multicolumn{1}{l}{Hous.} &
Ave.\tabularnewline
\midrule
\multirow{5}{0.28cm}{Prob.} &
L &
0.051 &
0.050 &
0.067 &
0.108 &
0.061 &
0.069 &
0.067\tabularnewline
 & PEP &
0.034 &
0.045 &
0.067 &
0.088 &
0.062 &
0.064 &
0.057\tabularnewline
 & SEP  &
0.034 &
0.045 &
0.067 &
0.088 &
0.062 &
0.064 &
0.057\tabularnewline
 & PPL  &
0.037 &
0.035 &
0.076 &
0.083 &
0.071 &
0.069 &
0.059\tabularnewline
 & SPL &
0.034 &
0.045 &
0.067 &
0.091 &
0.057 &
0.069 &
0.058\tabularnewline
\midrule
\multirow{5}{0.28cm}{Log.} &
L &
0.036 &
0.045 &
0.071 &
0.105 &
0.057 &
0.067 &
0.061\tabularnewline
 & PEP &
\textbf{0.251} &
0.045 &
0.071 &
0.083 &
0.062 &
0.070 &
\textbf{0.127}\tabularnewline
 & SEP &
0.034 &
0.045 &
0.067 &
0.083 &
0.062 &
0.070 &
0.057\tabularnewline
 & PPL  &
0.039 &
0.040 &
0.071 &
0.088 &
0.071 &
0.067 &
0.060\tabularnewline
 & SPL &
0.034 &
0.045 &
0.067 &
0.083 &
0.057 &
0.067 &
0.056\tabularnewline
\midrule
\multirow{5}{0.28cm}{NT} &
L &
- &
- &
- &
- &
- &
- &
-\tabularnewline
 & PEP &
\textbf{0.202} &
0.085 &
0.071 &
\textbf{0.384} &
\textbf{0.454} &
\textbf{0.124} &
\textbf{0.214}\tabularnewline
 & SEP &
0.034 &
0.045 &
0.067 &
0.086 &
0.062 &
0.069 &
0.058\tabularnewline
 & PPL  &
0.043 &
0.025 &
0.081 &
0.091 &
0.062 &
0.058 &
0.058\tabularnewline
 & SPL &
0.034 &
0.035 &
0.067 &
0.091 &
0.057 &
0.070 &
0.057\tabularnewline
\bottomrule
\end{tabular}
\end{table}

\section{Conclusions}

\label{sec:Conclusions}

We have proposed a novel method for classification using Gaussian
processes based on posterior linearisation. The proposed algorithm
consists of performing iterated statistical linear regressions with
respect to the current approximation to the posterior. An important
benefit compared to expectation propagation is that the proposed method
does not provide negative-definite covariance matrices by construction,
which implies that it does not need ad-hoc procedures to solve the
resulting problems. In addition, posterior linearisation has a local
convergence theorem. In the experimental results, PL and EP have
comparable performance except in the parallel implementations with
the noisy threshold likelihood where PL performs better than EP. 

\bibliographystyle{IEEEtran}
\bibliography{0E__Trabajo_Angel_Mis_articulos_Finished_PL_Gau___roccess_classification_Accepted_Referencias}

\cleardoublepage{}
\begin{center}
{\LARGE{}Supplementary material of ``Gaussian process classification
using posterior linearisation'' }
\par\end{center}{\LARGE \par}

In this supplementary material, we first provide the expressions for
the moments required in SLR for the probit, logit, and noisy-threshold
likelihood. We then analyse a synthetic example where EP, without
ad-hoc fixes, fails.

\section*{SLR for probit, logit and noisy-threshold likelihoods}

\label{subsec:SLR-for-different-likelihoods}

In this section, we provide the expressions for the SLR moments required
in (\ref{eq:A_plus})-(\ref{eq:Omega_plus}) for the probit, logit,
and noisy-threshold likelihoods. For the probit model, we have
\[
\mathrm{E}_{f_{i}}\left[\mathrm{E}\left[y_{i}|f_{i}\right]\right]=2\alpha-1,\,\mathrm{C}_{f_{i}}\left[\mathrm{E}\left[y_{i}|f_{i}\right]\right]+\mathrm{E}_{f_{i}}\left[\mathrm{C}\left[y_{i}|f_{i}\right]\right]=1-\alpha^{2},
\]
\[
\mathrm{C}_{f_{i}}\left[f_{i},\mathrm{E}\left[y_{i}|f_{i}\right]\right]=\frac{2P_{i}}{\sqrt{1+P_{i}}}\mathcal{N}\left(\frac{\overline{f}_{i}}{\sqrt{1+P_{i}}};0,1\right),
\]
where $\alpha=\Phi\left(\overline{f}_{i}/\sqrt{1+P_{i}}\right)$.
We have used Eq. (3.84) in \cite{Rasmussen_book06} to obtain the
last expression. 

For the noisy threshold model, we have 
\begin{align*}
\mathrm{E}_{f_{i}}\left[\mathrm{E}\left[y_{i}|f_{i}\right]\right] & =2\beta-1,\\
\mathrm{C}_{f_{i}}\left[\mathrm{E}\left[y_{i}|f_{i}\right]\right]+\mathrm{E}_{f_{i}}\left[\mathrm{C}\left[y_{i}|f_{i}\right]\right] & =1-\beta^{2},\\
\mathrm{C}_{f_{i}}\left[f_{i},\mathrm{E}\left[y_{i}|f_{i}\right]\right] & =2\left(1-2\epsilon\right)\sqrt{P_{i}}\\
 & \quad\times\mathcal{N}\left(\frac{\overline{f}_{i}}{\sqrt{P_{i}}};0,1\right),
\end{align*}
where $\beta=\left[1-\Phi\left(\overline{f}_{i}/\sqrt{P_{i}}\right)\right]\epsilon+\left(1-\epsilon\right)\Phi\left(\overline{f}_{i}/\sqrt{P_{i}}\right)$. 

For the logit model, the conditional moments are
\[
\mathrm{E}\left[y_{i}|f_{i}\right]=\frac{1-\exp\left(-f_{i}\right)}{1+\exp\left(-f_{i}\right)},\,\mathrm{C}\left[y_{i}|f_{i}\right]=1-\left(\frac{1-\exp\left(-f_{i}\right)}{1+\exp\left(-f_{i}\right)}\right)^{2}.
\]
The integrals $\mathrm{E}_{f_{i}}\left[\cdot\right]$ and $\mathrm{C}_{f_{i}}\left[\cdot\right]$
of these moments that are required in (\ref{eq:A_plus})-(\ref{eq:Omega_plus})
do not have closed-form expressions, but they can be approximated
using Gaussian quadrature rules/sigma-points \cite{Sarkka_book13,Julier04}.

\section*{Synthetic example where EP fails}

\label{subsec:Synthetic-example-where}

We consider a Gaussian prior over $\left(f_{1},f_{2}\right)$ with
mean $\left(-0.5,-3\right)$ and unit variances for both variables
with correlation 0.8. We also consider the noisy threshold likelihood,
see (\ref{eq:noisy_threshold_likelihood}), with $\epsilon=0.01$,
$y_{1}=1$, and $y_{2}=1$. 

We run EP first on the first variable and then on the second variable.
After the first round of updates, the cavity distribution over $f_{1}$
has a variance of -117.9, which stops the EP iterations. If we process
the measurements in the other order, we face the same problem. In
this case, the variance of the cavity distribution over $f_{1}$ is
-14.3. If we run EP in parallel \cite{Tolvanen14}, after the second
update, the cavity distribution over $f_{1}$ also has a variance
of -117.9. These negative variances are not due to numerical errors,
they are the result of the EP iterations. In particular, the problematic
part is the variance of the second variable, which increases after
its update, so the likelihood approximation has a negative-definite
variance. This creates problems at the next step of the iteration
\cite{Jylanki11}. 

\begin{figure}
\begin{centering}
\includegraphics[scale=0.6]{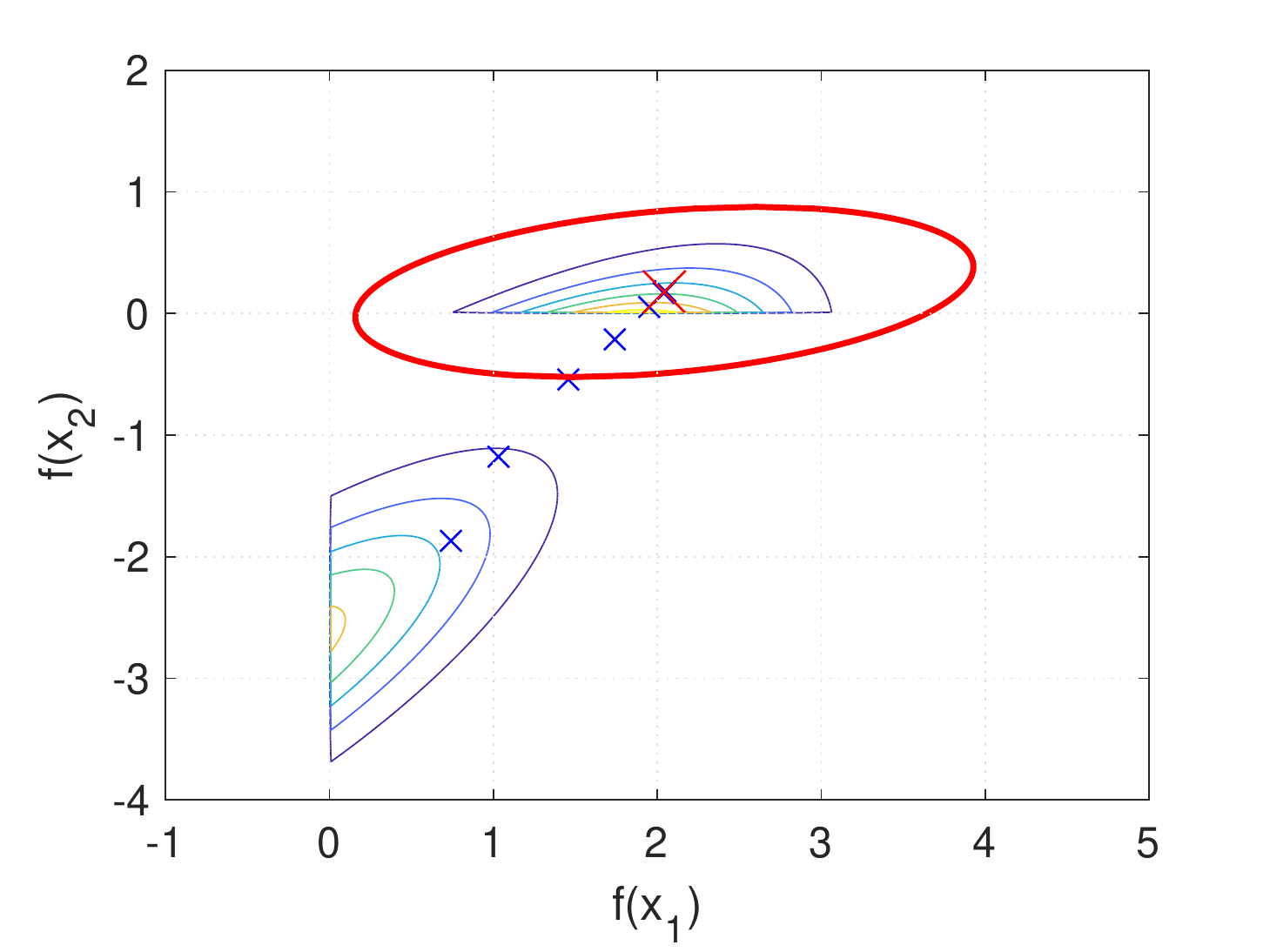}
\par\end{centering}
\caption{\label{fig:Examples}Contour plot of the posterior (Section \ref{subsec:Synthetic-example-where}).
PL posterior approximation: mean (red-cross) and the 3-$\sigma$ ellipse
(red line). Blue crosses represent the sequence of means of parallel
PL. The PL solution matches the mode with highest density: EP does
not provide a solution.}
\end{figure}

We show the contour plot of the posterior, which has been obtained
using a fine grid of points, and the PL solution in Figure \ref{fig:Examples}(a).
Both parallel and sequential implementations of PL converge to the
same solution and they match the mode of the posterior with highest
density, which is a reasonable Gaussian approximation to a bimodal
density. In the figure, we can also see the sequence of posterior
means provided by the parallel PL iterations. In 6 iterations, the
algorithm gets quite close to the fixed point. Laplace approximation
would not change the prior as the gradient of the likelihood is zero
almost everywhere. This example demonstrates that this type of situation
can be satisfactorily handled with PL, but not with EP, without fixes,
and Laplace methods. 

\end{document}